\newif\ifarxiv%
\newif\ifralfinal%
\arxivtrue%
\documentclass[letterpaper, 10 pt, conference]{ieeeconf}
\IEEEoverridecommandlockouts%
\pdfoutput=1%
\usepackage{amsmath,amssymb,amsfonts}%
\usepackage{algorithmic}%
\usepackage{algorithm}%
\usepackage{array}%
\usepackage{textcomp}%
\usepackage{stfloats}%
\usepackage{verbatim}%
\usepackage{graphicx}%
\graphicspath{{../figures/}}%
\DeclareGraphicsExtensions{.pdf,.jpeg,.png}%
\usepackage{cite}%
\usepackage{booktabs}%
\usepackage{hyperref}%
\usepackage[table,xcdraw]{xcolor}%
\usepackage{multirow}%
\usepackage{array}
\usepackage{tcolorbox}
\usepackage[super]{nth}%
\usepackage[top=60pt,bottom=50pt,left=50pt,right=50pt]{geometry}
\hypersetup{
  colorlinks=true,
  linkcolor=blue!80!black,
  urlcolor=blue,
  citecolor=green!50!black,
  linktocpage
}
\usepackage{caption}
\newcommand{\ocMCSMRT}[4]{\multicolumn{#1}{#2}{\multirow{#3}{12mm}{\centering{\B{#4}}}}}

\newcommand{\ocblbT}{\multicolumn{2}{|l|}{}}

\newcolumntype{M}[1]{>{\centering\arraybackslash}m{#1}}
\newcolumntype{L}[1]{>{\raggedright\arraybackslash}m{#1}}

\newcommand*{\B}[1]{\ifmmode\boldmath{#1}\else\textbf{#1}\fi}
\newcommand*{\I}[1]{\ifmmode\mathit{#1}\else\textit{#1}\fi}

\newcommand{\mdots}{\hbox to 1em{.\hss.\hss.\hss}}
\ifarxiv
    \usepackage{fancyhdr}
    \fancypagestyle{smallHeaders}{%
        \fancyhf{}
        \fancyhead[C]{ \fontsize{10}{10}\selectfont Preprint version: Accepted to IROS 2025. Final version available at (TBC)}
        \fancyfoot[C]{\thepage}
    }%
    
    \fancypagestyle{bigHeaders}{%
        \fancyhf{}
        \fancyhead[C]{ \fontsize{12}{12}\selectfont Preprint version: Accepted to IROS 2025. Final version available at (TBC)}
        
        \fancyfoot[C]{\thepage}
    }%
\fi

\begin{document}
    \title{\ifarxiv\LARGE\bf\fi
    Adversarial Attacks and Detection in Visual Place Recognition \\ for Safer Robot Navigation
    }%
    \author{%
        \parbox{\linewidth}{\centering
            Connor~Malone, %
            Owen~Claxton, %
            Iman~Shames, %
            Michael~Milford~\IEEEmembership{Senior Member,~IEEE}%
        }%
        %
        \thanks{This research is partially supported by an ARC Laureate Fellowship FL210100156 to M.Milford, the QUT Centre for Robotics, the Centre for Advanced Defence Research in Robotics and Autonomous Systems, and received funding from the Australian Government via grant AUSMURIB000001 associated with ONR MURI grant N00014-19-1-2571. The work of C.Malone was supported in part by an Australian Postgraduate Award. \textit{(Corresponding author: Connor Malone)}}%
        \thanks{C.Malone, O.Claxton, and M.Milford are with the QUT Centre for Robotics, School of Electrical Engineering and Robotics at the Queensland University of Technology, Brisbane, Australia (e-mail: \{cj.malone, o.claxton, michael.milford\}@qut.edu.au).}%
        \thanks{I.Shames is with the CIICADA Lab, College of Engineering, Computing and Cybernetics, at the Australian National University, Canberra, Australia (e-mail: iman.shames@anu.edu.au).}%
        %
    }

    \maketitle%
    \ifarxiv
        \thispagestyle{bigHeaders}
        \pagestyle{smallHeaders}
    \fi
    
    \begin{abstract}
        Stand-alone Visual Place Recognition (VPR) systems have little defence against a well-designed adversarial attack, which can lead to disastrous consequences when deployed for robot navigation. This paper extensively analyzes the effect of four adversarial attacks common in other perception tasks and four novel VPR-specific attacks on VPR localization performance. We then propose how to close the loop between VPR, an Adversarial Attack Detector (AAD), and active navigation decisions by demonstrating the performance benefit of simulated AADs in a novel experiment paradigm -- which we detail for the robotics community to use as a system framework. In the proposed experiment paradigm, we see the addition of AADs across a range of detection accuracies can improve performance over baseline; demonstrating a significant improvement -- such as a $\approx50\%$ reduction in the mean along-track localization error -- can be achieved with True Positive and False Positive detection rates of only 75\% and up to 25\% respectively. We examine a variety of metrics including: Along-Track Error, Percentage of Time Attacked, Percentage of Time in an `Unsafe' State, and Longest Continuous Time Under Attack. Expanding further on these results, we provide the first investigation into the efficacy of the Fast Gradient Sign Method (FGSM) adversarial attack for VPR.  The analysis in this work highlights the need for AADs in real-world systems for trustworthy navigation, and informs quantitative requirements for system design.

    \end{abstract}

    \section{Introduction}\label{sec:intro}

    \ifralfinal
\else
    Although
    \fi
    the impact of adversity in Visual Place Recognition (VPR) is widely understood, with state-of-the-art models offering increasing levels of robustness~\cite{VPRBench21,MixVPR23,anyloc,optimalsalad}, the effects of adversarial attacks remain under-explored. Adversarial attacks generally refer to perturbations made to signals or input data by adversaries, with the goal of forcing the output of a system to be incorrect~\cite{wang2022adversarial}. There has been a significant amount of work researching their effects on perception tasks such as image classification and object detection~\cite{wang2022adversarial, eykholt2018robust, saha2020role, liang2024securing, yang2020beyond}, yet they have not been widely investigated in the context of VPR.

    Adversarial attacks on perception systems vary depending on the level of access and information available to an attacker, including digital, physical-world, subtle, or overt attacks~\cite{wang2022adversarial}. In object detection tasks, this can result in misclassification or detection failure~\cite{liang2024securing}; for VPR, this can result in catastrophic localization failures that cause a complete inability to navigate an environment~\cite{ikram2022perceptual}. Consequently, it is important to foster further investigation into adversarial attacks in VPR to promote safer robot navigation.

    In this research, we perform one of the first investigations into how both naive and VPR-specific adversarial attacks affect localization metrics in VPR. We then explore how these effects can be mitigated with some form of attack detector in a novel active navigation strategy. Moving beyond this research foundation, we then provide the first study of how the Fast Gradient Sign Method (FGSM)~\cite{goodfellow2014explaining}, prominent in adversarial attack literature, affects VPR localization performance. In particular, we make the following contributions:

    \begin{figure}
        \centering
        \includegraphics[width=0.95\linewidth]{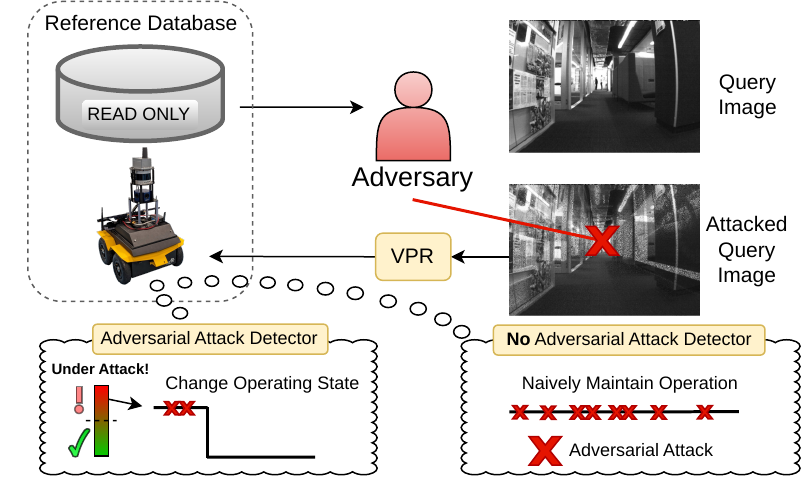}
        \caption{We investigate the under-explored effects of adversarial attacks in VPR and how they can be mitigated with Adversarial Attack Detectors (AADs). We show how black-box attacks can be more effectively designed in the context of VPR and propose how to close the loop between VPR, AAD, and active navigation behaviors for choosing safer navigation states.}
        \label{fig:main_diagram}
        \vspace*{-1.3\baselineskip}
    \end{figure}
    
    \begin{enumerate}
        \item We analyse how four common `black-box' type adversarial attacks used for other perception tasks and four novel, `smarter', VPR-specific adversarial attacks affect VPR performance over a range of ablations.
        \item We present an active navigation strategy for closing the loop between VPR, Adversarial Attack Detectors (AADs), and vehicle navigation.
        \item We introduce a novel experiment paradigm for assessing AADs, where performance is dependent on detection capability and corresponding active navigation choices.
        \item We extensively evaluate the proposed active navigation using AADs with a range of performance points on data from a variety of environments.
        \item We provide the first investigation into the efficacy of FGSM adversarial attacks for VPR.
    \end{enumerate}

    Our code and dataset features are publicly available\footnote{\label{githubrepo}\url{https://github.com/QVPR/aarapsiproject}}.
    
    \section{Background}\label{sec:background}

    \subsection{Visual Place Recognition}
        Visual Place Recognition (VPR) is widely researched as a method for approximate visual localization~\cite{Masone2021, Garg21_WYP}. It estimates position by comparing the similarity of a query image from the current place to a pre-generated `map' consisting of reference images and associated geographical metadata. The geographical position of the reference image most similar to the query becomes the localization estimate~\cite{Schubert23_VPR_Tutorial}. VPR can be used to navigate environments through teach-and-repeat type systems~\cite{Dom21TeachRepeat} or as a loop closure component in Simultaneous Localization And Mapping (SLAM).

        Current state-of-the-art VPR methods use various deep learning networks such as Convolutional Neural Networks (CNNs) and Vision Transformers (ViT) to extract abstract features from images and calculate similarities based on mathematical distances. Both traditional VPR networks such as NetVLAD~\cite{NetVLAD18} and AP-GeM~\cite{Revaud2019APGEM}, and recent approaches such as MixVPR~\cite{MixVPR23}, DinoV2 SALAD~\cite{optimalsalad}, and AnyLoc~\cite{anyloc} have achieved excellent performance on standard VPR benchmark datasets. Even under intentionally increased adversity, the most recent methods such as DinoV2 SALAD have shown excellent robustness~\cite{claxton2024improving}. However, there has been little investigation into their robustness to adversarial attacks.

    \subsection{Adversarial Attacks and Defences}
        Adversarial attacks on robot perception systems can be categorized in several different ways. They can be `white-box' attacks, where adversaries have access to the underlying perception system, or `black-box' attacks, where they have no access~\cite{wang2022adversarial}, and can be implemented digitally or in physical space, i.e., as either direct perturbations to digital data or as objects/changes in the physical world~\cite{liang2024securing}. Additionally, attacks can also be considered either misclassification attacks, where an attack confuses a chosen class/place with another, or evasion attacks, which prevent a system from detecting a chosen class/place~\cite{wang2022adversarial}. Misclassification attacks can be further separated into targeted or non-targeted attacks, depending on whether the adversary aims to confuse the chosen class/place with another specific class/place.

        Earlier literature investigating adversarial attacks for image classification and object detection showed that digital space attacks can be effective and covert to the human eye~\cite{szegedy2013intriguing}. More recent works demonstrate that effective attacks can also be implemented in the physical space~\cite{liang2024securing, ren2020adversarial}. Physical space attacks are generally more noticeable but can be easier for an adversary to deploy~\cite{eykholt2018robust}. However, in both cases, much of the research seems to optimize attacks in white-box scenarios.
        
        Whilst we don't explore specific defence strategies in this paper, there is a wide range of defences from object detection literature that could be applied to VPR. These can generally be separated into three categories: adversarial training, pre-processing, and detection~\cite{wang2022adversarial}. Pre-processing techniques include adding random noise, denoising, and compression~\cite{xie2017mitigating, liao2018defense, dziugaite2016study}. Whereas, current detection methods generally involve analyzing image properties such as contrast, homogeneity, entropy, or convolutional filter statistics~\cite{liang2024securing, li2017adversarial}.

        One of the few works for adversarial attacks in VPR investigates the problem in the context of loop closure in SLAM~\cite{ikram2022perceptual}. It shows a simple black-box attack made of a highly textured patch, which can be deployed digitally or physically, is highly effective at creating false loop closures in ORB-SLAM~\cite{mur2015orb}. The false loop closures distort the internal map, leading to catastrophic localization and navigation failures. The proposed defence for this attack was to provide any `friendly' robots with a database of sample attacked images, allowing for the rejection of false loop closures. To our knowledge, there is no in-depth analysis of how adversarial attacks and their design affects VPR-based localization.

    \subsection{Fast Gradient Sign Method}
    \label{subsec:litFGSM}
        The Fast Gradient Sign Method (FGSM)~\cite{goodfellow2014explaining} is a common adversarial attack in literature, typically used in classification tasks, based on the optimization process used for training neural networks. The FGSM attack uses gradients calculated during backpropagation to subtly augment an input image in a way which pushes it away from its true class in the feature space. Importantly, it was discovered that attacks generated to `fool' one network were also often effective against other networks as well~\cite{goodfellow2014explaining}. However, the performance of FGSM attacks in the context of VPR has not been explored.
        
        VPR networks are typically trained using a triplet contrastive loss rather than the classification loss which FGSM attacks are largely tested on~\cite{Schubert23_VPR_Tutorial}. Triplet losses are dependent on the sampling strategy for `positive' and `negative' samples, which makes the optimization target for attacks less straight-forward. This difference between generating attacks for the VPR and classification tasks is further exaggerated when the true location of a VPR query is unknown and triplet samples are based on a networks initial position estimate. Consequently, there is significant scope for investigations into the transferability and effectiveness of FGSM attacks for VPR. As an additional contribution in this work, we provide what we believe to be the first initial study into the efficacy of FGSM attacks for VPR.

    \section{Methodology}\label{sec:Approach}
    This work investigates how adversarial attacks affect VPR localization and how Adversarial Attack Detectors (AADs) can mitigate their effects. The following details the specific attacks for this investigation. The separate study into FGSM adversarial attacks is described later in Section~\ref{subsec:expFGSM}.

    \subsection{Adversarial Attacks}\label{subsec:AdvAttacks}
        Here, we explore some of the common attacks found in object detection literature as well as how these can be altered to specifically target VPR systems. We will focus on black-box, non-targeted, misclassification attacks, as this is representative of many real electronic/cyber-security scenarios where services/signals are disrupted or spoofed. That is, the attacks will aim to force incorrect VPR position estimates, not necessarily to confuse query images with any particular reference place. We use the term `image representation' to allow for the fact that attacks can be applied in the image space or in the feature space. We now define a selection of representative attacks used throughout this work, starting with some common black-box attacks found in existing literature:

        \begin{enumerate}
            \item \textbf{`Flat':} Takes a selection of pixels/locations in the query image representation and changes them to all have the same randomized value.
            \item \textbf{`Random':} Takes a selection of pixels/locations in the query image representation and randomizes their value.
        \end{enumerate}

        These types of attacks have been shown to be effective against camera and LiDAR perception systems when implemented as sensor attacks in autonomous vehicle contexts\cite{petit2015remote, yan2016can, cao2019adversarial, shin2017illusion}. Furthermore, \cite{ikram2022perceptual} showed that they can cause damaging loop closure errors in SLAM. However, we propose in the VPR scenario, these black-box attacks could be an even larger threat, as an adversary could potentially access and/or exploit the stored reference database which VPR relies on. In this work, we present the following VPR-specific attacks:

        \begin{enumerate}
        \setcounter{enumi}{2}
            \item \textbf{`Query'-based:} Takes a random selection of pixels/locations from a \textit{previous} query image representation and copies the values into corresponding locations from the \textit{current} query image representation.
            \label{attacktypes:query}
            \item \textbf{`Reference'-based:} Takes a random selection of pixels/locations from a random \textit{reference} image representation and copies the values into the corresponding locations in the \textit{query} image representation.
            \label{attacktypes:ref}
        \end{enumerate}

        These VPR-specific attacks have been designed to cause incorrect VPR position estimates through induced perceptual aliasing. They fall outside the classifications in current adversarial attack literature, which use the terms `black-box', `white-box', and sometimes `grey-box' to describe either access to the underlying perception system or an ability to duplicate it based on outputs. Instead, an adversary could achieve these attacks simply with access to query data - and the reference database for attack~\ref{attacktypes:ref}.
        
        With access to the reference database, it would be sensible for an adversary to attack references directly. However, given its importance, we propose here that the reference database would likely be stored as read-only data; therefore, there is a low chance of direct attack. However, the adversary could still disrupt VPR by reading values from the reference database and duplicating them in place of query data (Attack~\ref{attacktypes:ref}). Alternatively, if read access is not available but query data is accessible, similar attacks could be created by selecting random values from previously observed queries (Attack~\ref{attacktypes:query}).
        
        In this work, we explore how each of the adversarial attacks described above can be varied using two characteristics. Firstly, attacks can be generated as \textbf{Noise} randomly dispersed throughout an image representation, or in localized \textbf{Patch} attacks. Next, the percentage of data being attacked can be varied to control the trade off between `stealthiness' and effectiveness. That is, attacking more data is more likely to cause VPR errors but will be more noticeable, whereas attacking less data risks failing to cause errors.
        
\begin{figure}
            \centering
            \begin{tabular}{cccc}
                 \includegraphics[width=0.23\linewidth, height=0.17\linewidth]{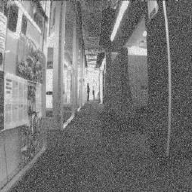} &  \includegraphics[width=0.23\linewidth, height=0.17\linewidth]{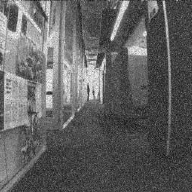} & \includegraphics[width=0.23\linewidth, height=0.17\linewidth]{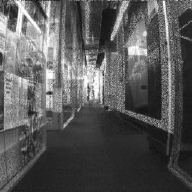} & \includegraphics[width=0.23\linewidth, height=0.17\linewidth]{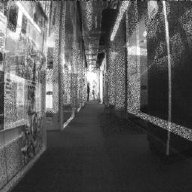} \\
                 \includegraphics[width=0.23\linewidth, height=0.17\linewidth]{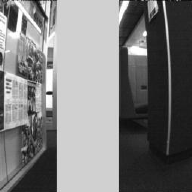} &  \includegraphics[width=0.23\linewidth, height=0.17\linewidth]{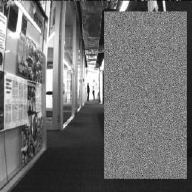} & \includegraphics[width=0.23\linewidth, height=0.17\linewidth]{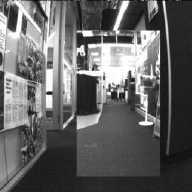} & \includegraphics[width=0.23\linewidth, height=0.17\linewidth]{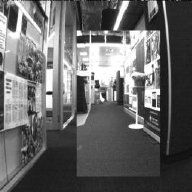}
            \end{tabular}
            \caption{Example Adversarial Attacks in the image space. \textbf{Top:} Noise-based attacks where random pixels are attacked. \textbf{Bottom:} Patch-based attacks, where whole patches are attacked. \textbf{Left to Right:} \textit{`Flat'}, \textit{`Random Noise'}, \textit{`Query'}-based attacks, and \textit{`Reference'}-based attacks.}
            \label{fig:attackegs}
            \vspace{-1.5\baselineskip}
        \end{figure}

        Figure~\ref{fig:attackegs} includes examples of the described attacks implemented digitally in image space, to aid in conceptualizing the attacks. However, in the experiments, we implement all attacks digitally and attack in the feature space, for simplicity.

    \subsection{Attack Detectors} \label{subsec:AttDetect}
    Given the prominence of adversarial attack/detection research in image classification and object detection, we propose similar attack detectors could be applied in VPR. With this in mind, this work instead focuses on investigating the overall effects of adversarial attacks on VPR-based localization and how they can be mitigated with some form of attack detector. We substitute any detection algorithm with simulated attack detectors covering a range of detection performance points.

        \setlength\tabcolsep{1.5pt}
        \setlength\tabcolsep{6pt}

         \begin{figure*}[ht]
            \centering
            \includegraphics[width=0.85\linewidth]{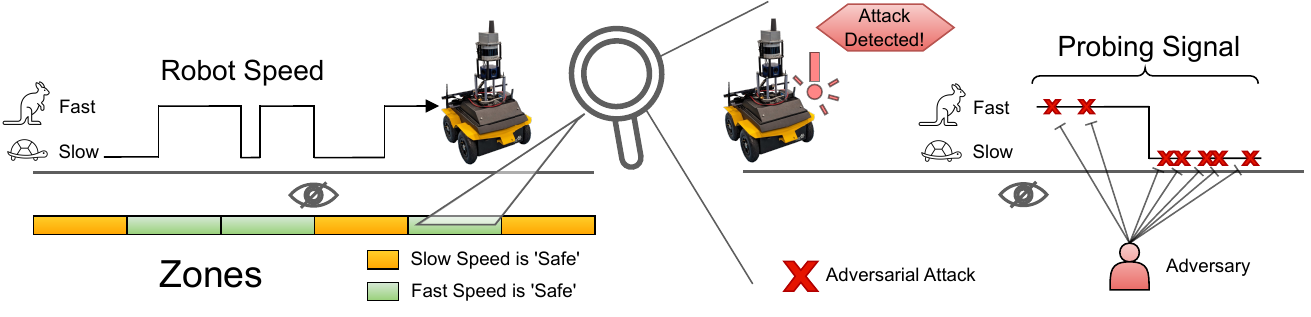}
            \caption{\textbf{Left:} Our Experimental design for evaluating the effect of adversarial attacks and AADs on VPR localization. Each zone is randomly allocated a speed that corresponds with being safest from attacks. If the robot is not driving at the `safe' speed, there is a high likelihood of an attack. \textbf{Right:} Our proposed active navigation behavior for closing the loop between VPR, AAD, and navigation decisions. The robot travels at each speed for a set duration and counts the number of attacks detected at each speed. The robot then drives onwards at the speed which incurred the least number of detected attacks.}
            \label{fig:expDes}
            \vspace*{-1.5\baselineskip}
        \end{figure*}
     \section{Experimental Procedure}\label{sec:ExpProcedure}

    In this section, we introduce our proposed experiment design for assessing: the effect of adversarial attacks in VPR; the ability of AADs to reduce their severity; and our novel active navigation strategy for closing the loop between VPR, AADs, and navigation decisions. We also discuss parameters and datasets.
    
    \subsection{Experiment Paradigm} \label{subsec:ExpParad}
        In this work, we focus on applications where a robot is navigating through a hostile environment without any knowledge of when or where in the environment it is most susceptible to adversarial attacks. We draw parallels here to real scenarios in electronic/cyber-security where sensor jamming/spoofing and Denial-of-Service attacks are real and can be unpredictable~\cite{yaugdereli2015study, petit2014potential, parkinson2017cyber}. We propose the following experimental scenario and design to evaluate the effects of both unnoticed and detected adversarial attacks on VPR.

        \subsubsection{The Scenario} \label{subsubsec:Scenario}
            A robot is traversing through a hostile environment where the likelihood of an adversarial attack is dependent on the robot's speed. Throughout the environment, the particular speed corresponding to a high likelihood of adversarial attack changes unknown to the robot. The adversarial attacks aim to damage the robot's ability to navigate by reducing the accuracy of VPR position estimates through the manipulation of query image representations. The goal of the robot is to navigate the environment whilst minimizing the number of attacks incurred and the localization error caused by the adversary. When equipped with an AAD, the robot can reject localization attempts while detecting attacks and vary its speed to reduce the likelihood of further attacks. We emphasize that the use of `speed' in this scenario is intended to be representative of different operating states such as, sensors, speed, or routes, and real situations where vehicle convoys or drones must change states to reduce susceptibility to threats with unknown precipitants.

        \subsubsection{Experiment Design} \label{subsubsec:ExpDesign}
            We evaluate the above scenario using the following design. The robot has two possible speeds from which it can select at any point: fast and slow. Throughout the environment, the `safe' speed corresponding with a lower likelihood of adversarial attacks will vary unknown to the robot between either the fast or slow speeds.

            The path through the environment is separated into a set of uniformly distributed `zones,' once again unknown to the robot. Each zone is randomly allocated with a `safe' speed (fast or slow), with a roughly equal number of zones allocated with each speed. We illustrate this setup in Figure~\ref{fig:expDes}.

            When the robot is traveling at the `safe' speed, there is only a 10\% chance of adversarial attacks; when the robot is \textbf{not} traveling at the `safe' speed, there is a 70\% chance of incurring adversarial attacks. When attacking, the adversary uses a randomly selected attack type from the list described in Section~\ref{subsec:AdvAttacks}. To account for randomness, we perform the experiment 100 times on each dataset with a new distribution of `safe' speed allocations across the zones each time. To maintain a fair comparison between the performance of VPR systems with and without AADs, each comparison method is evaluated on the same set of 100 randomized zones.

    \subsection{Defensive Measures} \label{subsec:AdvDef}

        Throughout the experiment, we evaluate different performance points for AADs. We set the performance baseline by first evaluating VPR methods with no ability to detect attacks. For the baseline VPR performance, the robot uses a set speed throughout the entire traverse; we measure the baseline performance at the fast speed. Since there is roughly an equal number of `safe' zones allocated to each speed, over the 100 randomized tests, the baseline performance at both the fast and slow speeds should be statistically equivalent.

        We also include a simple comparison where a new speed is randomly selected at the beginning of each `zone' to quantify the benefit AADs provide to VPR performance beyond the random chance of selecting the correct speed. We would like to emphasize that we include this comparison for diagnostic reasons; this is not a viable solution in real scenarios, as the robot does not know where the zones begin and end.

        For closing the loop between VPR, AADs, and navigation, we implement a two-part active navigation strategy to translate attack detections into a change in robot speed (Figure~\ref{fig:expDes} Right). The first step is accumulating a count of the number of attacks detected. Once the count exceeds a certain threshold, it triggers a probing behavior to determine the speed that results in the lowest likelihood of an adversarial attack. In the probing behavior, the robot operates at each speed for a set duration and counts the number of attacks detected at each speed in that time. The robot then selects to continue operating at the speed that resulted in fewer attacks during the probing time frame. Using this system, an AAD must maintain a sufficient number of correctly detected attacks without producing a significant number of falsely detected attacks.

        \subsection{Adversarial Attack Detectors} \label{subsec:AADs}
        We simulate the performance of AADs across a range of different possible attack detection accuracies. We parameterize these simulated AADs using a nominated rate of True Positive (TP) detection and False Positive (FP) detection, which we implement using randomly sampled binary values with these rates as probabilities. We evaluate AADs with TP values of $50\%, \ 60\%, \ 75\%, \ 85\%,$ and $95\%$, and corresponding FP values of $1-TP$, or, \ $50\%, \ 40\%, \ 25\%, \ 15\%,$ and $5\%$.

    \subsection{Metrics} \label{subsec:Metrics}

        We quantify the effect of adversarial attacks and detectors in the proposed experiment scenario using metrics such as: mean, median, and maximum along-track error of VPR position estimates; the total percentage of query images attacked throughout the traverse; the total percentage of query images in the traverse where the robot was operating at the `unsafe' speed; and the longest number of query images where the robot was continuously attacked.
        
        Additionally, we propose that when operating in a hostile environment, there is an increasing chance with every successful attack that the adversary will achieve a `Loss-of-Vehicle' (LoV) outcome, where the robot is no longer recoverable. In real scenarios, intermittent attacks can accumulate small errors over time and eventually cause navigation failures, or cause delays in time sensitive missions that result in failure. We propose that if more than $50\%$ of the traverse is attacked, this should correspond to a LoV event; we evaluate a continuous sweep of threshold values for completeness.

        \begin{figure}
            \centering
            \includegraphics[width=0.95\linewidth]{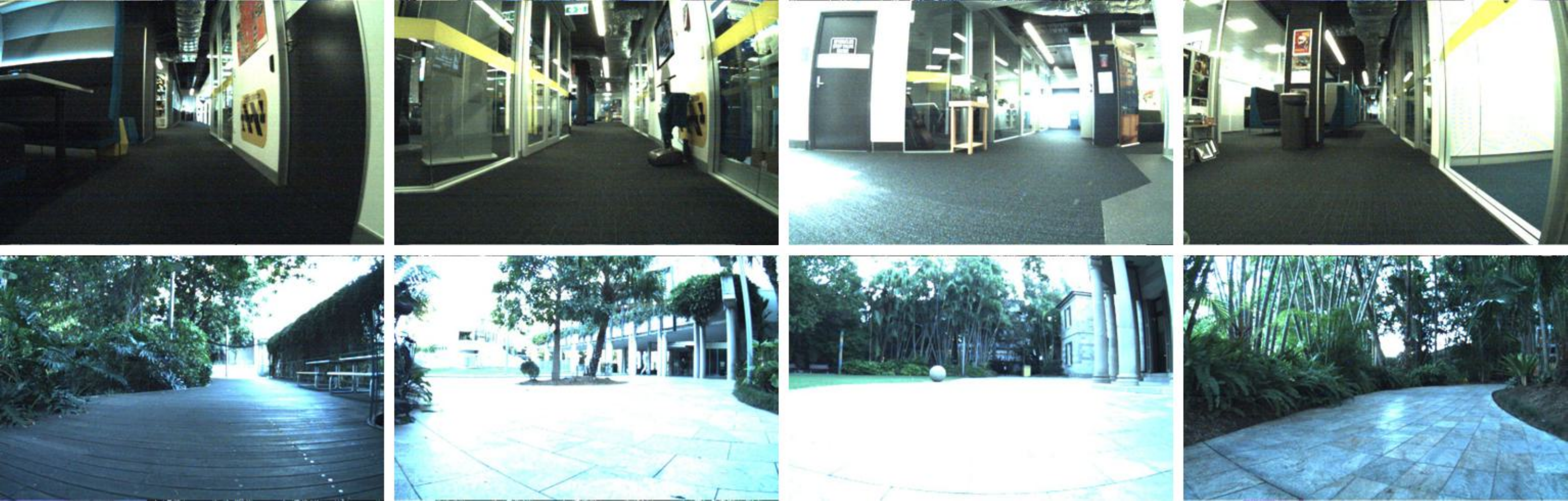}
            \caption{Example images from the datasets~\cite{claxton2024improving} used throughout the proposed experiment (\textbf{Top:} Office, \textbf{Bottom:} Campus). \textbf{Note:} We have selected images with no pedestrians captured to retain privacy. The original dataset images include pedestrians walking within the vicinity of the robot.}
            \label{fig:dataPics}
            \vspace{-1.5\baselineskip}
        \end{figure}

    \subsection{Datasets} \label{subsec:Datasets}

        We perform the described experiment across the QCR Office and Campus datasets, representative of both indoor and outdoor environments, collected and described in~\cite{claxton2024improving}. The QCR Office set contains images from a traverse around the QUT Centre for Robotics lab. The QCR Campus dataset contains images captured from around the QUT Gardens Point Campus. Each dataset contains a query and reference traverse captured at separate times. We include sample images from these datasets in Figure~\ref{fig:dataPics}.

    \subsection{VPR Descriptors}
    We evaluate results on multiple common and state-of-the-art VPR descriptors to demonstrate that the investigation into adversarial attacks in VPR is broadly applicable. For this work, we use AP-GeM~\cite{Revaud2019APGEM}, NetVLAD~\cite{NetVLAD18}, and DinoV2 SALAD~\cite{optimalsalad} (referred to as SALAD going forward). We compute VPR match distances using Euclidean distance. We emphasize, these descriptors are representative, this research does not aim to fault specific VPR descriptors. 

    \subsection{Adversarial Attack and Defence parameters}
    Adversarial attacks in the experimental scenario are formulated by selecting an attack at random from the list described in Section~\ref{subsec:AdvAttacks}, and then randomly varying the percentage of query data being attacked between $10\%$ and $50\%$. While it would be possible for an adversary to alter all data in this scenario, we assume stealth is important for the adversary to avoid countermeasures, and completely altering data would be a more easily detected. For this work, we choose to apply attacks in the VPR feature space \textit{after} features are extracted from images. In addition, we provide some analysis of the effectiveness of individual attack types.

    For defence against adversarial attacks, in Section~\ref{subsec:AdvDef}, we defined an active navigation strategy to vary speed and minimize attack frequency. To begin the described probing event, the accumulated number of attacks detected by the AAD must exceed some threshold value. We set this value to trigger a probing event after $10$ detected attacks, with a probe event length of $10$ query images per speed.

    \subsection{FGSM Attack Investigation}
    \label{subsec:expFGSM}
\begin{figure}
        \centering
        \begin{tabular}{cccc}
            \includegraphics[width=0.45\linewidth]{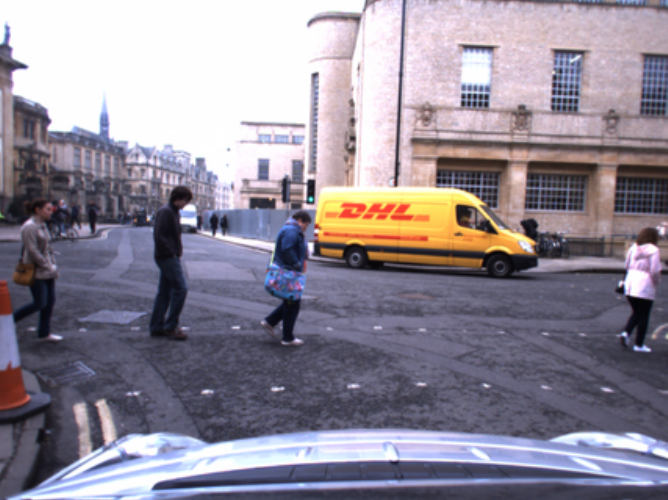} & 
            \includegraphics[width=0.45\linewidth]{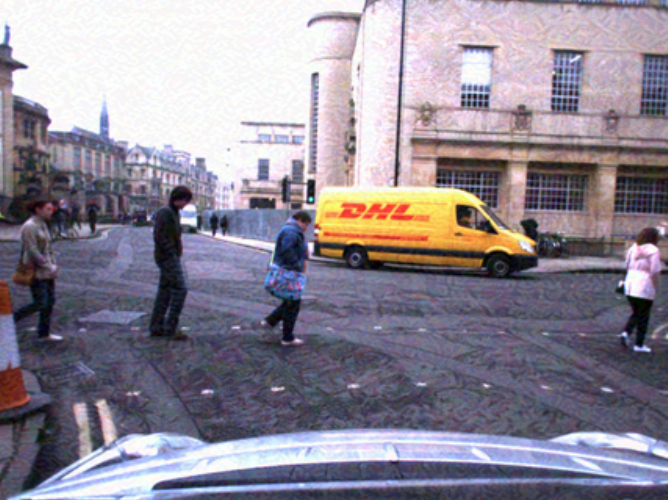}
        \end{tabular}
        \caption{Example images demonstrating subtle changes from an FGSM adversarial attack which cause VPR localization failure. \textbf{Left:} The original image. \textbf{Right:} The image after being attacked using FGSM.}
        \label{fig:attackFGSM}
        \vspace{-1.5\baselineskip}
    \end{figure}

    Separately to the described experiment, we provide an initial investigation into the use of FGSM adversarial attacks in VPR. For the purpose of this investigation, we utilise the Oxford RobotCar dataset~\cite{RobotCarDatasetIJRR}; using an `overcast' traverse with 3876 images as the query and a `sunny' traverse with 1:1 correspondence as the reference. To maintain the `black-box' context for this investigation, we use the MixVPR~\cite{MixVPR23} network to create adversarially attacked versions of each image. In real scenarios where ground truth information is unknown, these attacks would be generated with the objective of causing VPR matches far away from the initial position estimate. However, to observe the maximum potential effect these attacks could have on visual localization systems, we optimize attacks using the ground truth correspondence between query and reference images to generate positive and negative samples in the loss calculations. A sample of these attacks can be found in Figure~\ref{fig:attackFGSM}. In addition to the along-track localization error metrics described earlier, we also report the Recall@1, as calculated in~\cite{Schubert23_VPR_Tutorial}, using a ground truth tolerance of 3m.

    \setlength\tabcolsep{3pt}
    \setlength\tabcolsep{6pt}

    \section{Experimental Results}\label{sec:results}
Here, we present and discuss the results of the investigation into adversarial attacks and detection in VPR. We analyze how attack design affects localization error, and discuss observations from the proposed experiment. We evaluate using the metrics from Section~\ref{subsec:Metrics} to demonstrate both the severe impact adversarial attacks can have on VPR and the benefits of integrating an AAD into the navigation system.

    \subsection{Analysis of Adversarial Attacks}\label{subsec:advAttAna}
        
\begin{figure}
            \centering
            \includegraphics[width=0.95\linewidth]{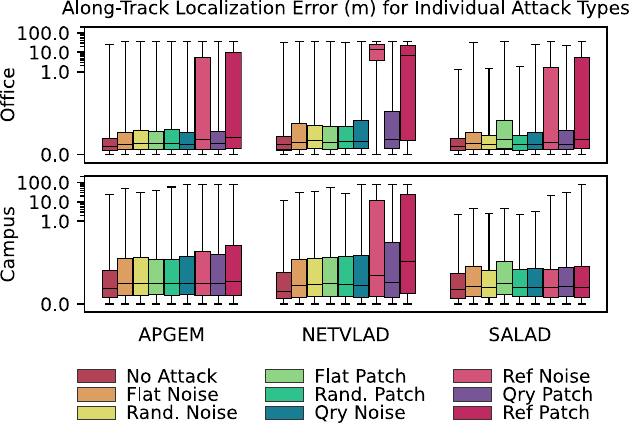}
            \caption{Box plots showing along-track localization error for attack types when attacking $40\%$ of data in each query. Trends remain consistent across VPR descriptors, with minimal differences between patch-based and noise-based attacks and that VPR is particularly vulnerable to reference-based attacks.}
            \label{fig:attTypAb}
            \vspace{-\baselineskip}
        \end{figure}

        Figure \ref{fig:attTypAb} demonstrates how the along-track localization error of each VPR descriptor is affected by the separate attack types described in Section~\ref{subsec:AdvAttacks}. It shows that all VPR descriptors are relatively robust to the `Flat' and `Random' attacks taken from existing literature, noting the split linear-logarithmic axis scale. This is likely a result of modern and state-of-the-art VPR descriptors using hyperdimensional feature vectors.
        
        Previous works investigating hyperdimensional computing in an image retrieval and VPR context have shown that the properties of hyperdimensional space dictate that any two randomly sampled vectors have a high likelihood of being nearly orthogonal~\cite{neubert2019introduction, Neubert2021Hyper}. The implication of this property being that the addition of a randomly sampled vector to the feature vector of a query has minimal impact on mathematical distances to reference feature vectors; making modern VPR descriptors relatively robust against `naive' black-box attacks.

        In this work, we also designed novel VPR-specific adversarial attacks (Attacks~\ref{attacktypes:query}-\ref{attacktypes:ref}). Figure~\ref{fig:attTypAb} shows that the attacks sampling values from previous queries for attacks were generally equally or slightly more effective than the Flat/Random attacks. This indicates that sampling values from random queries resembles random noise more than perceptual aliasing.

        In contrast, VPR-specific attacks sampling values from the reference feature vectors were the most effective for all VPR descriptors and often resulted in significant along-track localization errors. This indicates that the reference-based attacks have the desired effect of introducing perceptual aliasing. Across all attacks, there appears to be no substantial difference between `Noise' attacks and their `Patch'-based complement.
        
        Interestingly, Figure~\ref{fig:attTypAb} shows an increased effectiveness of the reference-based attacks on the Office dataset. This is likely a result of increased perceptual aliasing pre-existing in the dataset from repetitive structures and hallways often found in indoor environments. This pre-existing perceptual aliasing is then exaggerated by reference-based adversarial attacks, indicating VPR could be more susceptible to adversarial attacks in these types of environments.

\begin{figure}[ht]
            \centering
            \includegraphics[width=0.9\linewidth]{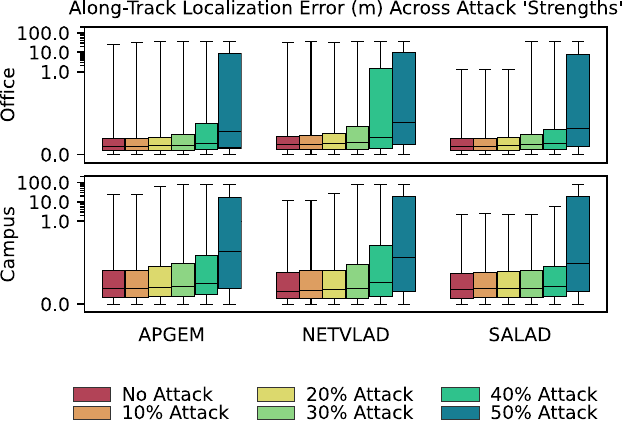}
            \caption{Box plots showing how the along-track localization error is correlated with varying the amount of query data affected by an adversarial attack. Attack types are randomized per query from Section~\ref{subsec:AdvAttacks}. It can be observed that there is a critical point from $40-50\%$ where error significantly increases.}
            \label{fig:attRange}
            \vspace{-0.5\baselineskip}
        \end{figure}

        We include Figure~\ref{fig:attRange} to demonstrate the effect of adversarial attacks as they are parameterized in the proposed experiment. It shows the change in along-track localization error as the amount of query data is increased, using attacks randomly chosen from the list in Section~\ref{subsec:AdvAttacks}. There is a clear spike in along-track localization error, and therefore the effectiveness of an attack, when at least $50\%$ of the query data is altered. This may be due to using mathematical distances for feature similarities; once queries lack a majority `original' data, similarities to reference features are more substantially impacted.

        While there are numerous more factors to investigate and analyze for adversarial attacks in VPR, we use these initial results to encourage further exploration of this topic.

        \begin{table}%
    \centering
    \caption{Along-Track localization error for VPR estimates in the experiment paradigm for assessing adversarial attack detection. At a TP rate of $75\%$, an AAD reduces mean along-track error by $\approx50\%$ for all VPR methods.}
    \vspace*{-0.5\baselineskip}
    \resizebox{\linewidth}{!}{
    \begin{tabular}{|lcL{10mm}|M{9mm}M{9mm}|M{10mm}M{10mm}M{10mm}M{10mm}M{10mm}|}
        \hline
        \multicolumn{10}{|c|}{\textbf{Office}} \\
        \hline
        &       &                   & \textbf{Baseline} & \textbf{Random} & \multicolumn{5}{c|}{\textbf{Adversarial Attack Detector}} \\
        &       &                   & \textbf{VPR}      & \textbf{Speed}  & \textbf{$50\%$ TP} & \textbf{$60\%$ TP} & \textbf{$75\%$ TP} & \textbf{$85\%$ TP} & \textbf{$95\%$ TP}\\
        \hline
        \ocMCSMRT{2}{|c|}{4}{AP-GeM} 
                & Mean   & 0.83m  & 0.81m  & 0.84m  & 0.59m  & 0.42m  & 0.35m  & 0.3m   \\
        \ocblbT & Median & 0.1m   & 0.1m   & 0.1m   & 0.1m   & 0.09m  & 0.09m  & 0.09m  \\
        \ocblbT & Max    & 34.67m & 34.67m & 34.67m & 34.67m & 34.67m & 34.67m & 34.67m \\
        \hline
        \ocMCSMRT{2}{|c|}{4}{NetVLAD} 
                & Mean   & 1.35m & 1.31m & 1.4m  & 1.02m & 0.69m & 0.57m & 0.49m  \\
        \ocblbT & Median & 0.13m & 0.13m & 0.13m & 0.12m & 0.12m & 0.12m & 0.12m  \\
        \ocblbT & Max    & 35.5m & 35.5m & 35.5m & 35.5m & 35.5m & 35.5m & 34.23m \\
        \hline
        \ocMCSMRT{2}{|c|}{4}{SALAD} 
                & Mean   & 0.6m   & 0.58m  & 0.57m  & 0.44m  & 0.25m  & 0.2m   & 0.16m  \\
        \ocblbT & Median & 0.11m  & 0.11m  & 0.11m  & 0.11m  & 0.1m   & 0.1m   & 0.1m   \\
        \ocblbT & Max    & 35.32m & 35.32m & 35.32m & 35.32m & 35.32m & 35.32m & 32.45m \\
        \hline
        \multicolumn{10}{|c|}{\textbf{Campus}} \\
        \hline
        &       &                   & \textbf{Baseline} & \textbf{Random} & \multicolumn{5}{c|}{\textbf{Adversarial Attack Detector}} \\ 
        &       &                   & \textbf{VPR}      & \textbf{Speed}  & \textbf{$50\%$ TP}  & \textbf{$60\%$ TP}  & \textbf{$75\%$ TP}  & \textbf{$85\%$ TP}  & \textbf{$95\%$ TP}\\
        \hline
        \ocMCSMRT{2}{|c|}{4}{AP-GeM} 
                & Mean   & 1.03m & 1.02m & 1.01m & 0.73m & 0.49m & 0.41m & 0.35m \\
        \ocblbT & Median & 0.2m  & 0.2m  & 0.2m  & 0.2m  & 0.19m & 0.19m & 0.19m \\
        \ocblbT & Max    & 81.7m & 81.7m & 81.7m & 81.7m & 81.7m & 81.7m & 81.7m \\
        \hline
        \ocMCSMRT{2}{|c|}{4}{NetVLAD} 
                & Mean   & 1.44m  & 1.4m   & 1.41m  & 1.01m  & 0.57m  & 0.45m  & 0.35m  \\
        \ocblbT & Median & 0.18m  & 0.18m  & 0.18m  & 0.17m  & 0.16m  & 0.16m  & 0.16m  \\
        \ocblbT & Max    & 81.59m & 81.59m & 81.59m & 81.59m & 81.59m & 81.59m & 81.59m \\
        \hline
        \ocMCSMRT{2}{|c|}{4}{SALAD} 
                & Mean   & 0.65m & 0.66m & 0.67m & 0.52m & 0.38m & 0.32m & 0.29m  \\
        \ocblbT & Median & 0.19m & 0.19m & 0.19m & 0.18m & 0.18m & 0.18m & 0.18m  \\
        \ocblbT & Max    & 80.8m & 80.8m & 80.8m & 80.8m & 80.8m & 80.8m & 78.61m \\
        \hline
    \end{tabular}}
    \vspace{-1.5\baselineskip}
    \label{table:combined}
\end{table}

    \subsection{Attack Detection Effect on Localization and Navigation} \label{subsec:genRes}

\begin{figure*}[ht]
            \centering
            \begin{tabular}{cc}
                 \includegraphics[width=0.45\linewidth]{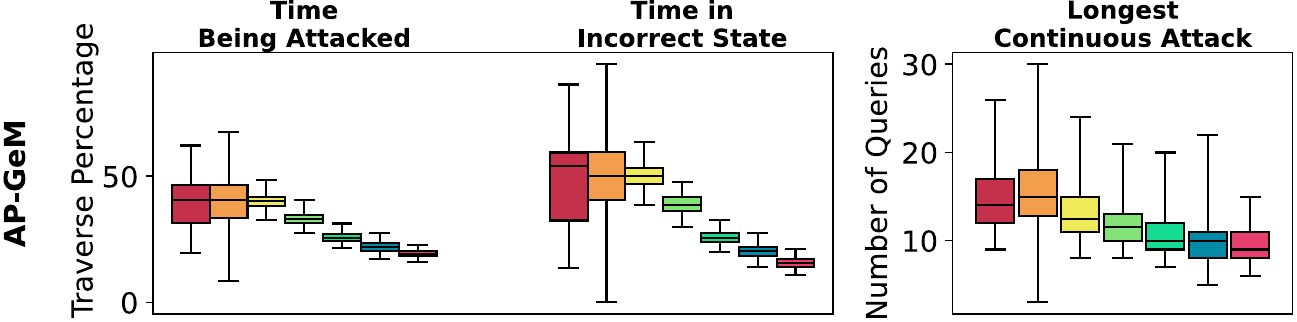} & \includegraphics[width=0.45\linewidth]{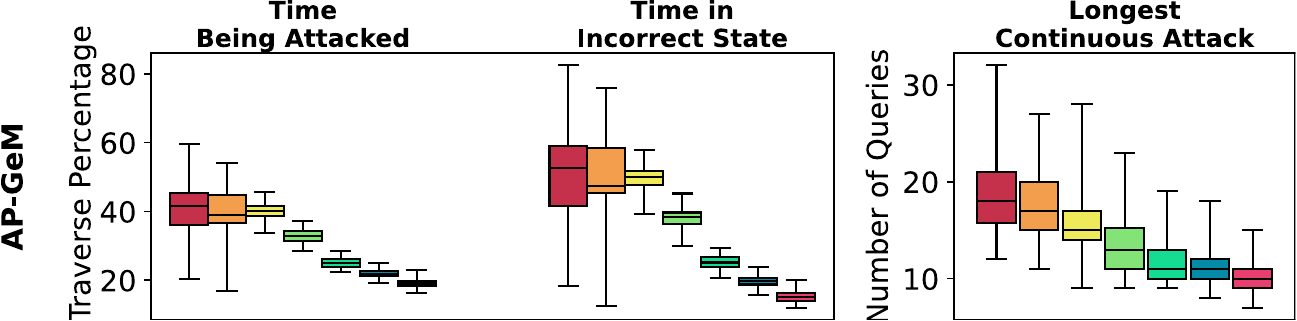} \\
                 \includegraphics[width=0.45\linewidth]{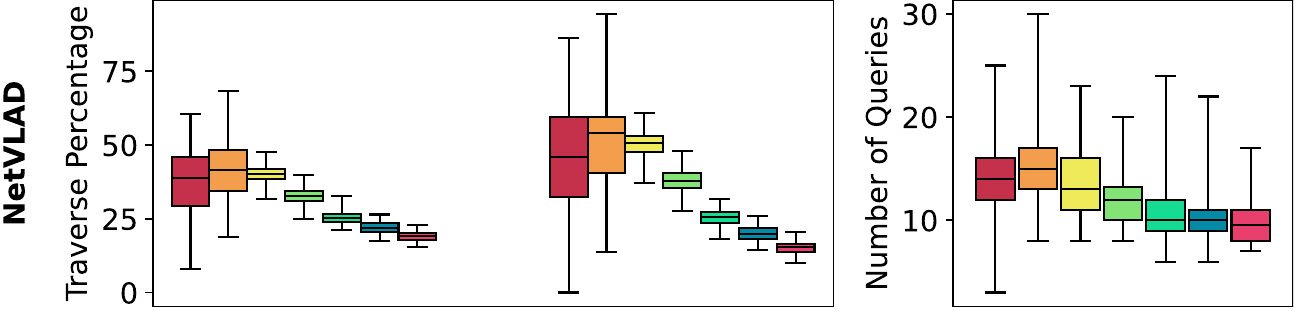} & \includegraphics[width=0.45\linewidth]{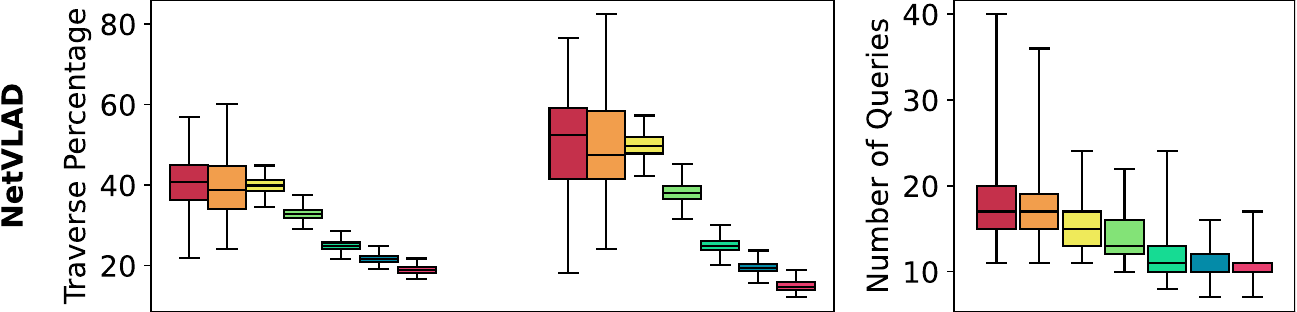} \\
                 \includegraphics[width=0.45\linewidth]{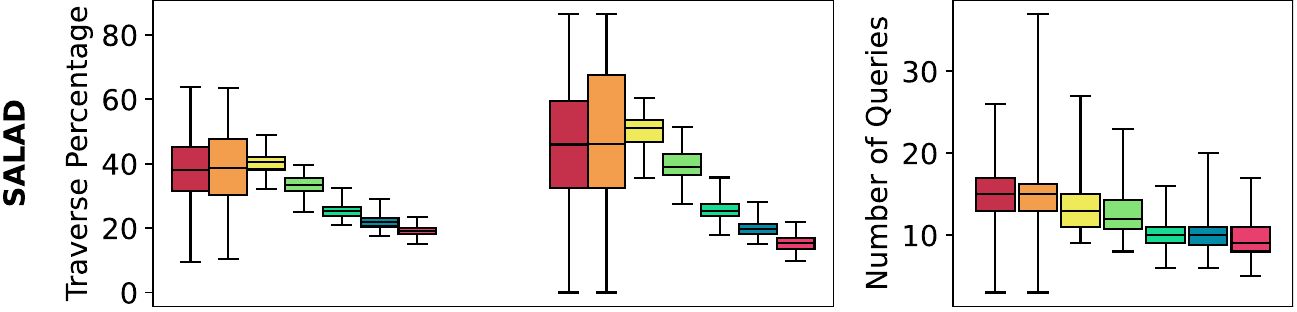} & \includegraphics[width=0.45\linewidth]{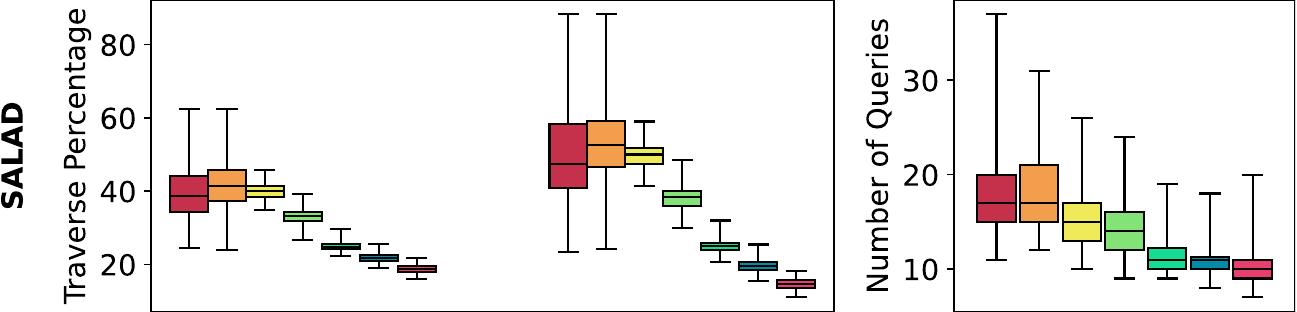} \\
                 \multicolumn{2}{c}{\includegraphics[width=0.8\linewidth]{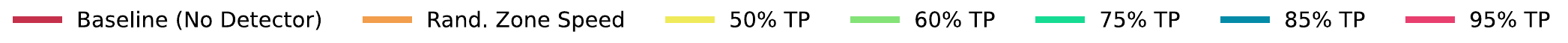}}
            \end{tabular}
            \caption{\textbf{Left:} Office. \textbf{Right:} Campus. Box plots visualizing the attack-based metrics across our proposed experiment both with and without an AAD. In general, the random speed selection for each zone is approximately equivalent to selecting one speed across the traverse - likely a result of the 50/50 distribution of zones with fast and slow `safe' speeds. AADs increasingly improve these metrics as accuracy increases, with a performance jump at a TP rate of $75\%$.}
            \label{fig:attBoxPlots}
            \vspace{-\baselineskip}
        \end{figure*}

        Here, we discuss the results in Table~\ref{table:combined} and Figure~\ref{fig:attBoxPlots} from the proposed experimental paradigm, with attacks of $10-50\%$. Table~\ref{table:combined} shows that the addition of an AAD has little effect on median and maximum along-track localization error values, but makes a clear and significant improvement to the mean error. At a True Positive rate of just $75\%$, the AAD typically reduces the mean error by $\approx50\%$ over the baseline with no detector. These trends hold across all VPR descriptors and both environments.

        At $50\%$ TP, the AAD does not improve along-track localization error, however, Figure~\ref{fig:attBoxPlots} shows a reduction in variation of the attack-based metrics (time under attack, and time traveling at the incorrect speed), compared to the baseline and random zone speed methods. This may be desirable, however, the corresponding high False Positive rate causes more `probing events' which results in unnecessary changes in speed, making it not ideal in real scenarios.

        The attack-based metrics in Figure~\ref{fig:attBoxPlots} are a visualization of the risk of navigating a hostile environment. The more a robot is attacked, the higher chance there is of failing the objective. Logically, there is an inverse correlation where time under attack and time in the \textit{incorrect} state decreases as the AAD accuracy increases. At $75\%$ TP, there is a jump in performance improvements followed by diminishing returns for all VPR descriptors and in both environments. Interestingly, even though the chance of attack for each query is independent and randomly sampled, the longest continuous attack also decreases as the AAD detection accuracy increases.

    \subsection{Loss of Vehicle (LoV) Ablation} \label{subsec:lovAbl}
\begin{figure}
            \centering
            \includegraphics[width=\linewidth]{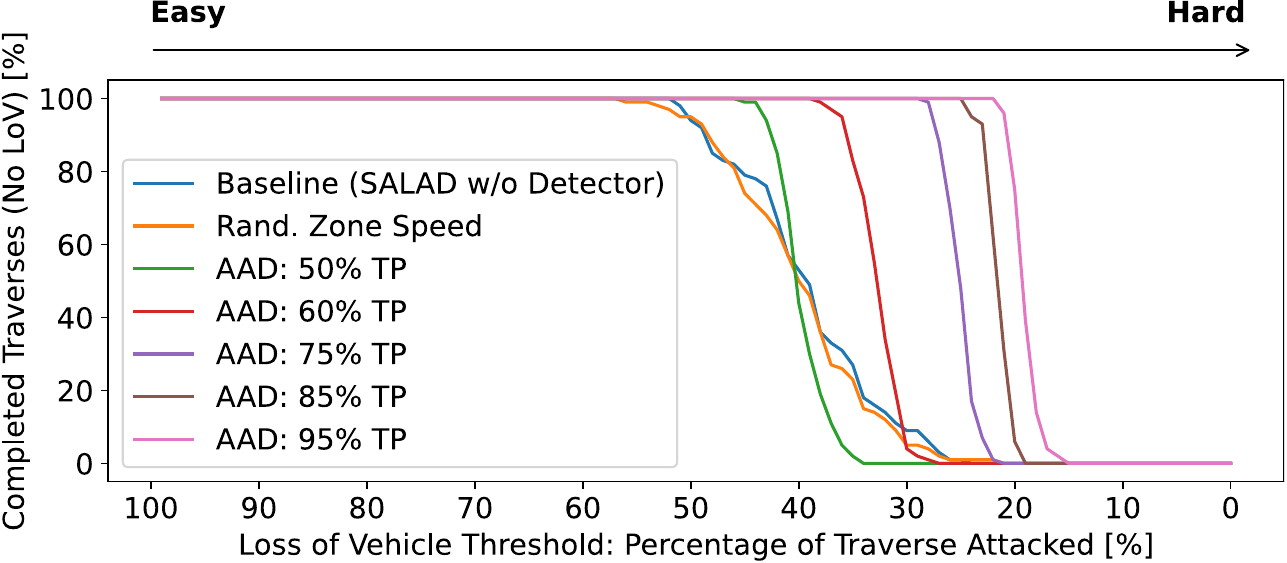}
            \caption{Ablation of the threshold of a traverse that must be attacked before an LoV event occurs - using SALAD on the QCR Campus Dataset. At $75\%$ TP rate, an AAD avoids any LoV where the baseline completes no traverses.}
            \label{fig:lovAbl}
            \vspace{-\baselineskip}
        \end{figure}

        In Section~\ref{subsec:Metrics}, we proposed that once a certain percentage of queries was successfully attacked, a Loss of Vehicle (LoV) event would occur. We reiterate here that this metric is grounded in real scenarios where navigation errors can accumulate to create drift or delays can cause failure of a time-sensitive mission. Figure \ref{fig:lovAbl} shows an ablation on the threshold selection for the LoV metric using SALAD on the QCR Campus dataset. Evidently, all detectors show a steep gradient from $0\%$ of traverses completed to $100\%$, each with a different and narrow response range.
        
        At the nominated threshold of $50\%$, all detectors complete 100\% of traverses without any LoV. The baseline performance with no detector and the random speed comparison generally completed between $75\%$ and $95\%$ of traverses for each VPR descriptor and both environments at the same threshold. However, Figure~\ref{fig:lovAbl} shows there is a significant increase in the benefits of AADs with respect to LoV when the threshold is reduced. At $33\%$, the traverses completed by the baseline drops to ${10-20\%}$, whereas all detectors with TP rates above $60\%$ typically remain at $100\%$ of traverses completed.
    
    \subsection{FGSM Attacks in Visual Place Recognition}
    \label{subsec:fgsm}
    
    In this section, we discuss results from the investigation into FGSM attacks for VPR; emphasizing that this is not an exhaustive study and aims to encourage further research. Table~\ref{tab:fgsm} shows the along-track localization error (ATE) and Recall@1 of each VPR method for both the original images and the FGSM attacked images from the Oxford RobotCar datasets. The first column of results clearly shows a significant decrease in performance for the VPR method used to generate attacks, MixVPR in this case. While the three other VPR methods also decrease in performance, the effect is much less than for MixVPR. This indicates that FGSM attacks may not be as transferable for VPR as they are in the classification task. However, the use of gradients to generate attacks may make generated features seem more natural and therefore make these attacks harder to detect. Overall, it is clear VPR has at least some vulnerability to FGSM adversarial attacks and therefore their effect on visual localization should be more thoroughly investigated.

    \begin{table}%
    \centering
    \begin{tabular}{|l|c|ccc|}
    \hline
       \textbf{ Method $\rightarrow$} & \textbf{MixVPR} & \textbf{AP-GeM} & \textbf{NetVLAD} & \textbf{SALAD} \\
    \hline
        & \multicolumn{4}{c|}{\textbf{Original Images}} \\
    \hline
        Mean ATE & 2.86m & 50.8m & 103.6m & 1.80m \\
        Median ATE & 1.0m & 1.0m & 1.0m & 1.0m \\
        Max ATE & 2993m & 3172m & 3584m & 1793m \\
        Recall@1 & 86.2 & 71.1 & 68.0 & 84.8 \\
    \hline
        & \multicolumn{4}{c|}{\textbf{FGSM Attacked Images}} \\
    \hline
        Mean ATE & 353.5m & 118.0m & 338.6m & 2.43m \\
        Median ATE & 8.0m & 2.0m & 3.0m & 1.0m \\
        Max ATE & 3742m & 3172m & 3758m & 315m \\
        Recall@1 & 24.5 & 60.2 & 45.6 & 67.9 \\
    \hline
    \end{tabular}
    \caption{Localization performance under adversarial attack using FGSM~\cite{goodfellow2014explaining} on the Oxford RobotCar dataset~\cite{RobotCarDatasetIJRR} with an `Overcast' query and a `Sunny' reference. Attacks are generated using MixVPR~\cite{MixVPR23} and transferred to other VPR methods to maintain the `black-box' nature of this work. The table reports along-track localization error and Recall@1 as calculated in~\cite{Schubert23_VPR_Tutorial}.}
    \label{tab:fgsm}
    \vspace{-2\baselineskip}
\end{table}

    \section{Conclusion}
Adversarial attacks are a real and possible threat when a robot is navigating hostile environments. This paper investigated the under-explored impact of adversarial attacks on VPR and how they effect a robot's ability to localize and navigate. Using both established black-box attacks, and novel VPR-specific attacks, we showed that state-of-the-art VPR descriptors are generally robust to random noise but susceptible to attacks exploiting perceptual aliasing. Expanding on this research foundation, we demonstrated that FGSM adversarial attacks may not be as transferable as for classification networks, but VPR methods still show some vulnerability, indicating a need for further investigations.

We evaluated AADs across a sweep of detection accuracies and found a TP rate of only $75\%$ reduced mean along track error by $\approx50\%$ for all VPR descriptors across both environments and number of incurred attacks. Importantly, responses from VPR descriptors to adversarial attacks and detectors were consistent for all results. Future work can further this research by using existing literature from object detection and/or VPR integrity~\cite{Carson22, Carson23, claxton2024improving, zaffar2024estimation} to design AADs with these operational frameworks and guidelines in mind.

The presented experimental paradigm demonstrates the need for adversarial attack detection in a simplified scenario, but its applicability extends further than this. Rather than varying operating speed, a vehicle/drone could switch between various sensors, or any number of operating states to respond to threats in search or delivery missions. In general, this work aimed to contribute to, and highlight the need for, deeper analysis in the under-explored field of adversarial attacks in VPR for safer robot navigation.

    \bibliographystyle{IEEEtran}
    \bibliography{IEEEabrv,ref.bib}

\begin{thebibliography}{10}
\providecommand{\url}[1]{#1}
\csname url@samestyle\endcsname
\providecommand{\newblock}{\relax}
\providecommand{\bibinfo}[2]{#2}
\providecommand{\BIBentrySTDinterwordspacing}{\spaceskip=0pt\relax}
\providecommand{\BIBentryALTinterwordstretchfactor}{4}
\providecommand{\BIBentryALTinterwordspacing}{\spaceskip=\fontdimen2\font plus
\BIBentryALTinterwordstretchfactor\fontdimen3\font minus
  \fontdimen4\font\relax}
\providecommand{\BIBforeignlanguage}[2]{{%
\expandafter\ifx\csname l@#1\endcsname\relax
\typeout{** WARNING: IEEEtran.bst: No hyphenation pattern has been}%
\typeout{** loaded for the language `#1'. Using the pattern for}%
\typeout{** the default language instead.}%
\else
\language=\csname l@#1\endcsname
\fi
#2}}
\providecommand{\BIBdecl}{\relax}
\BIBdecl

\bibitem{VPRBench21}
M.~Zaffar and et~al., ``{VPR-Bench}: An open-source visual place recognition
  evaluation framework with quantifiable viewpoint and appearance change,''
  \emph{International Journal of Computer Vision}, May 2021.

\bibitem{MixVPR23}
A.~Ali-Bey, B.~Chaib-Draa, and P.~Giguére, ``Mixvpr: Feature mixing for visual
  place recognition,'' in \emph{IEEE/CVF Winter Conference on Applications of
  Computer Vision}, 2023, pp. 2997--3006.

\bibitem{anyloc}
N.~Keetha and et~al., ``Anyloc: Towards universal visual place recognition,''
  \emph{arXiv preprint arXiv:2308.00688}, 2023.

\bibitem{optimalsalad}
S.~Izquierdo and J.~Civera, ``Optimal transport aggregation for visual place
  recognition,'' in \emph{IEEE/CVF Conference on Computer Vision and Pattern
  Recognition}, 2024, pp. 17\,658--17\,668.

\bibitem{wang2022adversarial}
J.~Wang, C.~Wang, Q.~Lin, C.~Luo, C.~Wu, and J.~Li, ``Adversarial attacks and
  defenses in deep learning for image recognition: A survey,''
  \emph{Neurocomputing}, vol. 514, pp. 162--181, 2022.

\bibitem{eykholt2018robust}
K.~Eykholt and et~al., ``Robust physical-world attacks on deep learning visual
  classification,'' in \emph{Proceedings of the IEEE conference on computer
  vision and pattern recognition}, 2018, pp. 1625--1634.

\bibitem{saha2020role}
A.~Saha, A.~Subramanya, K.~Patil, and H.~Pirsiavash, ``Role of spatial context
  in adversarial robustness for object detection,'' in \emph{Proceedings of the
  IEEE/CVF Conference on Computer Vision and Pattern Recognition Workshops},
  2020, pp. 784--785.

\bibitem{liang2024securing}
J.~Liang, R.~Yi, J.~Chen, Y.~Nie, and H.~Zhang, ``Securing autonomous vehicles
  visual perception: Adversarial patch attack and defense schemes with
  experimetal validations,'' \emph{IEEE Transactions on Intelligent Vehicles},
  2024.

\bibitem{yang2020beyond}
K.~Yang, T.~Tsai, H.~Yu, T.-Y. Ho, and Y.~Jin, ``Beyond digital domain: Fooling
  deep learning based recognition system in physical world,'' in
  \emph{Proceedings of the AAAI Conference on Artificial Intelligence},
  vol.~34, no.~01, 2020, pp. 1088--1095.

\bibitem{ikram2022perceptual}
M.~H. Ikram, S.~Khaliq, M.~L. Anjum, and W.~Hussain, ``Perceptual aliasing++:
  Adversarial attack for visual slam front-end and back-end,'' \emph{IEEE
  Robotics and Automation Letters}, vol.~7, no.~2, 2022.

\bibitem{goodfellow2014explaining}
I.~J. Goodfellow, J.~Shlens, and C.~Szegedy, ``Explaining and harnessing
  adversarial examples,'' \emph{arXiv preprint arXiv:1412.6572}, 2014.

\bibitem{Masone2021}
C.~Masone and B.~Caputo, ``A survey on deep visual place recognition,''
  \emph{IEEE Access}, vol.~9, pp. 19\,516--19\,547, 2021.

\bibitem{Garg21_WYP}
S.~Garg, T.~Fischer, and M.~Milford, ``Where is your place, visual place
  recognition?'' in \emph{13th International Joint Conference on Artificial
  Intelligence}, Z.-H. Zhou, Ed., 2021, pp. 4416--4425.

\bibitem{Schubert23_VPR_Tutorial}
S.~Schubert, P.~Neubert, S.~Garg, M.~Milford, and T.~Fischer, ``Visual place
  recognition: A tutorial,'' \emph{IEEE Robotics \& Automation Magazine}, pp.
  2--16, 2023.

\bibitem{Dom21TeachRepeat}
D.~Dall’Osto, T.~Fischer, and M.~Milford, ``Fast and robust bio-inspired
  teach and repeat navigation,'' in \emph{2021 IEEE/RSJ International
  Conference on Intelligent Robots and Systems (IROS)}, 2021.

\bibitem{NetVLAD18}
R.~Arandjelović, P.~Gronat, A.~Torii, T.~Pajdla, and J.~Sivic, ``Netvlad: Cnn
  architecture for weakly supervised place recognition,'' \emph{IEEE
  Transactions on Pattern Analysis and Machine Intelligence}, vol.~40, no.~6,
  pp. 1437--1451, 2018.

\bibitem{Revaud2019APGEM}
J.~Revaud, J.~Almaz{\'a}n, R.~S. Rezende, and C.~R.~d. Souza, ``Learning with
  average precision: Training image retrieval with a listwise loss,'' in
  \emph{IEEE/CVF International Conference on Computer Vision}, 2019.

\bibitem{claxton2024improving}
O.~Claxton and et~al., ``Improving visual place recognition based robot
  navigation by verifying localization estimates,'' \emph{IEEE Robotics and
  Automation Letters}, 2024.

\bibitem{szegedy2013intriguing}
C.~Szegedy and et~al., ``Intriguing properties of neural networks,''
  \emph{arXiv preprint arXiv:1312.6199}, 2013.

\bibitem{ren2020adversarial}
H.~Ren and T.~Huang, ``Adversarial example attacks in the physical world,'' in
  \emph{Machine Learning for Cyber Security}.\hskip 1em plus 0.5em minus
  0.4em\relax Springer, 2020, pp. 572--582.

\bibitem{xie2017mitigating}
C.~Xie, J.~Wang, Z.~Zhang, Z.~Ren, and A.~Yuille, ``Mitigating adversarial
  effects through randomization,'' \emph{arXiv:1711.01991}, 2017.

\bibitem{liao2018defense}
F.~Liao, M.~Liang, Y.~Dong, T.~Pang, X.~Hu, and J.~Zhu, ``Defense against
  adversarial attacks using high-level representation guided denoiser,'' in
  \emph{Proceedings of the IEEE conference on computer vision and pattern
  recognition}, 2018, pp. 1778--1787.

\bibitem{dziugaite2016study}
G.~K. Dziugaite, Z.~Ghahramani, and D.~M. Roy, ``A study of the effect of jpg
  compression on adversarial images,'' \emph{arXiv preprint arXiv:1608.00853},
  2016.

\bibitem{li2017adversarial}
X.~Li and F.~Li, ``Adversarial examples detection in deep networks with
  convolutional filter statistics,'' in \emph{Proceedings of the IEEE
  international conference on computer vision}, 2017, pp. 5764--5772.

\bibitem{mur2015orb}
R.~Mur-Artal, J.~M.~M. Montiel, and J.~D. Tardos, ``Orb-slam: a versatile and
  accurate monocular slam system,'' \emph{IEEE transactions on robotics},
  vol.~31, no.~5, pp. 1147--1163, 2015.

\bibitem{petit2015remote}
J.~Petit, B.~Stottelaar, M.~Feiri, and F.~Kargl, ``Remote attacks on automated
  vehicles sensors: Experiments on camera and lidar,'' \emph{Black Hat Europe},
  vol.~11, no. 2015, p. 995, 2015.

\bibitem{yan2016can}
C.~Yan, W.~Xu, and J.~Liu, ``Can you trust autonomous vehicles: Contactless
  attacks against sensors of self-driving vehicle,'' \emph{Def Con}, vol.~24,
  no.~8, p. 109, 2016.

\bibitem{cao2019adversarial}
Y.~Cao and et~al., ``Adversarial sensor attack on lidar-based perception in
  autonomous driving,'' in \emph{ACM SIGSAC conference on computer and
  communications security}, 2019, pp. 2267--2281.

\bibitem{shin2017illusion}
H.~Shin, D.~Kim, Y.~Kwon, and Y.~Kim, ``Illusion and dazzle: Adversarial
  optical channel exploits against lidars for automotive applications,'' in
  \emph{Cryptographic Hardware and Embedded Systems--CHES 2017: 19th
  International Conference, Taipei, Taiwan, September 25-28, 2017,
  Proceedings}.\hskip 1em plus 0.5em minus 0.4em\relax Springer, 2017, pp.
  445--467.

\bibitem{yaugdereli2015study}
E.~Ya{\u{g}}dereli, C.~Gemci, and A.~Z. Akta{\c{s}}, ``A study on
  cyber-security of autonomous and unmanned vehicles,'' \emph{The Journal of
  Defense Modeling and Simulation}, vol.~12, no.~4, pp. 369--381, 2015.

\bibitem{petit2014potential}
J.~Petit and S.~E. Shladover, ``Potential cyberattacks on automated vehicles,''
  \emph{IEEE Transactions on Intelligent transportation systems}, vol.~16,
  no.~2, pp. 546--556, 2014.

\bibitem{parkinson2017cyber}
S.~Parkinson, P.~Ward, K.~Wilson, and J.~Miller, ``Cyber threats facing
  autonomous and connected vehicles: Future challenges,'' \emph{IEEE
  transactions on intelligent transportation systems}, vol.~18, no.~11, pp.
  2898--2915, 2017.

\bibitem{RobotCarDatasetIJRR}
W.~Maddern, G.~Pascoe, C.~Linegar, and P.~Newman, ``{1 Year, 1000km: The Oxford
  RobotCar Dataset},'' \emph{The International Journal of Robotics Research
  (IJRR)}, vol.~36, pp. 3--15, 2017.

\bibitem{neubert2019introduction}
P.~Neubert, S.~Schubert, and P.~Protzel, ``An introduction to hyperdimensional
  computing for robotics,'' \emph{KI-K{\"u}nstliche Intelligenz}, vol.~33,
  no.~4, pp. 319--330, 2019.

\bibitem{Neubert2021Hyper}
P.~Neubert and S.~Schubert, ``Hyperdimensional computing as a framework for
  systematic aggregation of image descriptors,'' in \emph{2021 IEEE/CVF
  Conference on Computer Vision and Pattern Recognition (CVPR)}, 2021, pp.
  16\,933--16\,942.

\bibitem{Carson22}
H.~Carson, J.~J. Ford, and M.~Milford, ``Predicting to improve: Integrity
  measures for assessing visual localization performance,'' \emph{IEEE Robotics
  and Automation Letters}, vol.~7, no.~4, pp. 9627--9634, 2022.

\bibitem{Carson23}
------, ``Unsupervised quality prediction for improved single-frame and
  weighted sequential visual place recognition,'' in \emph{2023 IEEE
  International Conference on Robotics and Automation (ICRA)}, 2023.

\bibitem{zaffar2024estimation}
M.~Zaffar, L.~Nan, and J.~F. Kooij, ``On the estimation of image-matching
  uncertainty in visual place recognition,'' in \emph{Proceedings of the
  IEEE/CVF Conference on Computer Vision and Pattern Recognition}, 2024, pp.
  17\,743--17\,753.

\end{thebibliography}

    \vfill
\end{document}